\documentclass{jot}

\usepackage{multirow}
\usepackage{amsmath,amssymb,amsfonts}
\usepackage{graphicx}
\usepackage{textcomp}
\usepackage{xcolor}
\usepackage{todonotes}
\usepackage{multirow}
\usepackage{booktabs}
\usepackage{listings}
\usepackage{threeparttable}
\usepackage{tcolorbox}
\usepackage{pdfpages}

\usepackage{eso-pic}
\AtBeginDocument{%
  \AddToShipoutPictureBG*{%
    \AtPageUpperLeft{%
      \raisebox{-3.5cm}{% 调整这个数值对准空白框
        \hspace{1.5cm}% 调整水平位置
        \parbox{0.85\textwidth}{%
          \small\itshape\color{red} This is a Preprint. The final version is published in the Journal of Object Technology. 
        }%
      }%
    }%
  }%
}

\jotdetails{
    volume=vv,      % volume
    number=nn,       % number or issue
    articleno=aa,   % article number, eg a1 for research articles, e for editorials
    year=yyyy,      % year
    license=ccbyncnd    % choose from ccby, ccbynd, ccbyncnd
}

\newtcolorbox{promptbox}{
colback=white,
colframe=cyan!40!black,
rounded corners,
boxsep=2pt,
left=4pt,
right=4pt,
top=3pt,
bottom=3pt
}

\newtcolorbox{answerbox}{
colback=white,
colframe=gray!70!black,
rounded corners,
boxsep=2pt,
left=4pt,
right=4pt,
top=3pt,
bottom=3pt
}

\newcommand{\updated}[1]{\textcolor{black}{#1}}
\newcommand{\revised}[1]{\textcolor{black}{#1}}

\articletype{regular}

\title{Leveraging LLMs for Grammar Adaptation: A Study on Metamodel-Grammar Co-Evolution}

\author[$\ast$]{Weixing Zhang}
\author[$\ast$]{Bowen Jiang}
\author[$\ast$]{Rahul Sharma}
\author[$\dagger$]{Regina Hebig}
\author[$\ddagger$,$\S$]{Daniel Strüber}

\affil[$\ast$]{Karlsruhe Institute of Techonology, Germany}
\affil[$\dagger$]{Universität Rostock, Germany}
\affil[$\ddagger$]{Chalmers University of Technology and University of Gothenburg, Sweden}
\affil[$\S$]{Radboud University, The Netherlands}


\runningtitle{LLM-based Metamodel-Grammar Co-Evolution}  

\runningauthor{Zhang \textit{et al.}}

\begin{abstract}
In model-driven engineering, metamodel evolution leads to the need to adapt corresponding grammars to maintain consistency, which typically requires tedious manual work. Existing rule-based methods can achieve partial automation but have limitations when handling complex grammar scenarios. This paper proposes a Large Language Model-based approach that automatically applies adaptations to new grammars after evolution by learning grammar adaptations from previous versions. We evaluated this approach on six real-world Xtext domain-specific languages, using four DSLs as a training set to develop prompting strategies, two DSLs as a test set for validation, and conducting a longitudinal case study on QVTo. The evaluation used three Large Language Models (Claude Sonnet 4.5, ChatGPT 5.1, Gemini 3) and measured grammar adaptation quality from three dimensions: grammar rule-level adaptation consistency, output similarity, and metamodel conformance. Results show that on the test set, all three LLMs achieved 100\% adaptation consistency and output similarity, while the rule-based approach achieved only 84.21\% on DOT and 62.50\% on Xcore. In the QVTo longitudinal study, the LLM-based approach successfully reused learned adaptations across all three evolution steps without manual grammar editing, while the rule-based approach required manual adjustments in two of three transitions. However, on large-scale grammars (EAST-ADL, 297 rules), LLMs' adaptation consistency was far below 90\%. This study demonstrates the advantages of LLM-based approaches in handling complex grammar scenarios, while revealing their limitations in large-scale grammar adaptation.
\end{abstract}

\acknowledgment{
This work was funded by the Deutsche Forschungsgemeinschaft (DFG, German Research Foundation) - SFB 1608 - 501798263 and supported by the pilot program Core Informatics at KIT (KiKIT) of the Helmholtz Association (HGF). It was partially funded by Vetenskapsrådet, grant 2021-04881\_VR.
}

\begin{document}
% \includepdf[pages=-]{response_letter.pdf}

\maketitle
\urlstyle{rm}

\section{Introduction}
\label{sec:intro}
\begin{figure*}[tb]
  \centering
  \includegraphics[width=0.8\linewidth]{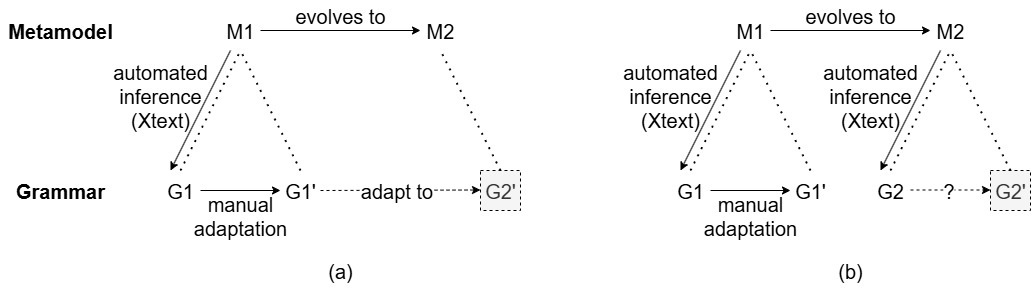}
  \caption{\updated{Challenges in metamodel-grammar co-evolution: (a) Direct evolution scenario where language engineers manually propagate metamodel changes from $M_1$ to $M_2$ into the grammar evolution from $G_1'$ to $G_2'$ after metamodel evolution; (b) Regeneration scenario where $G_2$ is generated from $M_2$ and the adaptations from $G_1$ to $G_1'$ need to be reapplied to obtain $G_2'$.}
  % The grammar $G_1$ generated from the metamodel $M_1$ was manually adapted to $G_1'$. When the metamodel evolved to $M_2$, a new grammar $G_2$ is generated in order for the grammar to continue to conform to the metamodel. However, the manual adaptations made in the previous version of the grammar could not be automatically replayed in $G_2$.
  }
  \label{fig:problem}
\end{figure*}

% \todo[inline]{Weixing to @all: Here lacks many citations, and I will add them later. Or just revise it, because this section is rather rough so far.}
% Domain-Specific Languages (DSLs) play an increasingly important role in software engineering, providing highly specialized expressiveness for specific domain problems, thereby improving development efficiency and reducing error rates. As a powerful DSL development framework, Eclipse Xtext~\cite{XtextHomepage} has been widely applied in various scenarios, supporting automatic grammar definition generation from domain metamodels, which provides convenience for language engineers.
In model-driven engineering (MDE), evolution is a long-standing core challenge~\cite{paige2016evolving}. When a metamodel changes, various software artifacts that depend on it—including model instances, model transformation rules, constraint definitions, etc.—need to be adapted accordingly to maintain consistency~\cite{MEYERS20111223, hebig2016approaches}. This multi-artifact co-evolution problem has been widely studied, with existing work mainly focusing on co-evolution between models and metamodels~\cite{kessentini2019automated, cicchetti2008meta, cicchetti2008automating}, as well as co-evolution between model transformations and metamodels~\cite{garcia2012model, kessentini2018automated}. However, in the development and maintenance of domain-specific languages (DSLs), there exists another type of co-evolution scenario: when the abstract syntax (usually represented as a metamodel) evolves, the grammar definition that specifies the concrete syntax also needs to evolve accordingly~\cite{rath2010synchronization, tolvanen2025framework}.
This metamodel-driven approach is \revised{particularly} common in scenarios where the metamodel is the central development artifact in a larger ecosystem, e.g., \textit{blended modeling} \cite{ciccozzi2019blended}, where several concrete syntaxes for the same underlying metamodel exist and evolve in parallel.
When DSL development adopts the metamodel-driven approach, it is necessary to support co-evolution between metamodels and grammar definitions.
\updated{One widely used language framework, Xtext~\cite{XtextHomepage, erdweg2013state}, supports this metamodel-centric DSL development. In Xtext, the metamodel defines the abstract syntax of the language, while the grammar defines the concrete syntax, and the two adhere to each other. When the metamodel exists first, we can generate grammar from it~\cite{bettini2016implementing}. The generated grammar usually has to be manually adapted, since it may include verbose formatting, e.g, redundant parantheses and keywords, leading to the modification of $G_1$ to derive $G_1'$ in Figure~\ref{fig:problem}(a))~\cite{bettini2016implementing}. However, when the metamodel evolves (from $M_1$ to $M_2$), language engineers need to adapt $G_1'$ according to the changes in the metamodel (as shown in $G_1'$ to $G_2'$ in Figure~\ref{fig:problem}(a)). This adaptation is particularly laborious and error-prone, since the language engineers manually have to create new rules for newly added elements, based on their understanding of the metamodel changes, an involved task especially in case of complex changes.
Xtext provides another alternative scenario (i.e., scenario (b) in Figure~\ref{fig:problem}), where, after the metamodel evolves, language engineers directly generate grammar $G_2$ from the evolved metamodel $M_2$ and perform the adaptation on $G_2$ that occurred in $G_1$ to $G_1'$. However, this introduces repetitive work and is equally prone to errors.}
% This manual adaptation faces challenges during metamodel evolution (Figure~\ref{fig:problem}): when the metamodel is updated, the manual adaptations made in the previous version cannot be automatically applied to the new version of the grammar, causing language engineers to need to repeat the adaptations. This repetitive manual effort is not only time-consuming but also prone to introducing inconsistencies during repetition.

To address these challenges in scenario (b), previous research proposed the GrammarTransformer tool~\cite{zhang2024supporting}, which contains 60 predefined grammar adaptation rules that enable automated grammar adaptation through configuration. This approach reduces the complexity of manual modification to some extent; however, when adaptation requirements vary significantly across grammar rules, the configuration process can still become cumbersome. Subsequent work further attempted to automatically extract adaptation configurations by comparing the generated grammar with the target grammar, but its adaptation accuracy still needs improvement~\cite{zhang2023automated}.

To propose an improved solution to the above challenges, this study explores using Large Language Models (LLMs) to automate the grammar adaptation process. LLMs have demonstrated strong capabilities in code understanding~\cite{wang2023codet5+} and transformation tasks~\cite{cummins2024don}, suggesting potential for learning adaptations from examples. 
A recent systematic mapping of LLM applications in MDE confirms broad interest in this intersection, while also revealing that tasks such as DSL Engineering remain marginal, with only three of 86 surveyed studies addressing them~\cite{zhang2026llm4mde}.
Unlike rule-based methods, our approach does not rely on predefined adaptation rules, but instead investigates whether LLMs can automatically adapt new grammars by learning from historical adaptations. 
Specifically, we provide the LLM with the grammar generated from the original metamodel ($G_1$) and its manually adapted version ($G'_1$), as well as the new grammar generated from the evolved metamodel ($G_2$), to evaluate whether the LLM can identify the adaptations from $G_1$ to $G'_1$ and apply the same adaptations to $G_2$ to generate $G'_2$. The generated $G'_2$ needs to both maintain consistency with the evolved metamodel and reproduce the manual adaptations from the previous version.

To evaluate the LLM-based approaches, we propose the following research questions:\\
\textbf{RQ1}: To what extent can LLMs learn and apply grammar adaptations from \revised{a prior-version grammar pair (G1, G'1)?}\\
% historical examples?\\
\textbf{RQ2}: \revised{To what extent} can LLM-based approaches \revised{reuse learned adaptations across consecutive evolution steps} %support multi-version language evolution 
with minimal human intervention?\\
\textbf{RQ3}: What are the strengths and limitations of LLM-based approaches compared to rule-based methods for grammar adaptation?

We systematically evaluated the proposed approach 
% To answer the RQs, we applied the LLM-based approaches
on six real-world DSLs from prior work~\cite{zhang2023automated}, using a training set (EAST-ADL~\cite{eatop-bitbucket}, BibTeX~\cite{BibTeX}, Xenia~\cite{Xenia}, SML~\cite{smlRepo}) to develop prompting strategies and validating on a test set (DOT~\cite{DotXtext}, Xcore~\cite{EclipseXcore}). Additionally, we conducted a longitudinal case study on QVTo spanning four official versions. We used three representative current LLMs (Claude Sonnet 4.5~\cite{claudesonnet4.5}, ChatGPT 5.1~\cite{chatgpt5.1}, Gemini 3~\cite{gemini3}) in the evaluation. 
% Results show that on the test set DSLs, all three LLMs achieved 100\% \revised{grammar rule-level} adaptation consistency (RAC) and 100\% output similarity, while the rule-based approach achieved 84.21\% RAC and 87.50\% output similarity on DOT, and 62.50\% RAC and 70.00\% output similarity on Xcore. 
Results show that on the test set DSLs, all three LLMs outperformed the rule-based approach in adaptation consistency and output similarity. 
In the QVTo longitudinal study, the LLM-based approach 
% achieved 100\% RAC with 
\revised{successfully reused learned adaptations across all three evolution steps %with 100\% RAC and 
without manual grammar editing,}
% zero human intervention 
% across all three evolution steps, 
while the rule-based approach required configuration adjustments in two of the three transitions. 
However, on large-scale grammars such as EAST-ADL with 297 rules, LLMs encountered challenges, exhibiting systematic omission of identified adaptation operations. %with RAC values far below 90\%. 
This study demonstrates that LLM-based approaches can reduce language engineers' workload, particularly in handling complex grammar scenarios that are difficult for rule-based methods to address (such as syntactic predicates, predicated assignments, and order-insensitive attribute combinations), while also revealing the limitations of current LLMs in large-scale grammar adaptation tasks, pointing out directions that require further exploration in the field of metamodel-grammar co-evolution.

% The main contributions of this research include:
% \begin{itemize}
%     \item Systematically evaluating the capabilities of LLMs in Xtext grammar adaptation, verifying the feasibility of automatically adapting generated grammars after evolution through learning historical adaptations.
%     \item Exploring the application potential of LLMs in the field of grammar style customization, analyzing their ability to customize DSL grammar styles according to user preferences or domain characteristics.
%     \item Proposing an experimental framework combining grammar adaptation and style customization, providing new ideas for Xtext grammar optimization.
% \end{itemize}

% Section~\ref{sec:background} of this paper introduces necessary background; Section~\ref{sec:approach} describes our method in detail; Section~\ref{sec:evaluation} presents the experimental design and results; Section~\ref{sec:discussion} discusses the findings, limitations, and threats to validity; Section~\ref{sec:relatedwork} reviews related work; and finally, Section~\ref{sec:conclusion} summarizes the main findings and contributions of this paper.

The remainder of this paper is structured as follows: Section~\ref{sec:background} provides background; Sections~\ref{sec:approach}–\ref{sec:evaluation} describe the approach and evaluation; Sections~\ref{sec:discussion}–\ref{sec:conclusion} present discussion, related work, and conclusion.

\section{Background}
\label{sec:background}
% \todo[inline]{Assigned to @Bowen}
\subsection{Metamodel-Driven Development and Grammar adaptation}
% \todo[inline]{Weixing: the first paragraph should be simplified, and the term "transformation" should be introduced.}
Xtext is a widely-used language development framework~\cite{erdweg2013state} that supports both grammar-driven and metamodel-driven development workflows~\cite{bettini2016implementing, paige2014tutorial}. 
In the grammar-driven development workflow, language engineers primarily define the concrete syntax by editing the grammar, from which the corresponding metamodel is automatically generated.
In contrast, in the metamodel-driven development workflow, language engineers first create a metamodel to represent domain knowledge and capture the abstract syntax of the language. From this metamodel, the grammar can be automatically generated.

As mentioned in Section~\ref{sec:intro}, 
% the generated grammar typically follow a default structure, which
% % This grammar 
% \revised{contains} multiple grammar rules, and each grammar rule contains multiple attributes. These grammar rules and their attributes correspond one-to-one with the metaclasses and properties in the metamodel
the generated grammar \revised{contains} multiple grammar rules and attributes that correspond one-to-one with the metaclasses and properties in the metamodel, sharing the same names~\cite{bettini2016implementing}. 
% The generated grammar always has a default format, and its main characteristics include that 
\updated{The grammar rules of generated grammar} always begin with the same name as the metaclass and often contain a keyword with the same name as the metaclass, while attributes are enclosed in curly braces~\cite{bettini2016implementing}. Similarly, attribute keywords and names are the same as the corresponding properties defined in the metamodel. 
% Language engineers often adapt the generated grammar according to their own needs.  

\begin{lstlisting}[basicstyle=\footnotesize\ttfamily, breaklines=true, label=lst:example, caption={A grammar rule \texttt{Mission} before adaptation.}]
Mission returns Mission:
	'Mission'
	'{'
		'shortName' shortName=Identifier
		('category' category=Identifier)?
		('uuid' uuid=String0)?
		('name' name=String0)?
		('ownedComment' '{' ownedComment+=Comment ( "," ownedComment+=Comment)* '}' )?
	'}';
\end{lstlisting}

\begin{lstlisting}[basicstyle=\footnotesize\ttfamily, breaklines=true, label=lst:example_1, caption={The grammar rule \texttt{Mission} after adaptation.}]
Mission returns Mission:
    'Mission'
         shortName=Identifier
    ('{'
        ('category' category=Identifier ';')?
        ('uuid' uuid=UUID ';')?
        ('name' name=Identifier ';')?
        (  ownedComment+=Comment (  ownedComment+=Comment)*  )?
    '}')?;
\end{lstlisting}

\textbf{\textit{Grammar adaptation}} refers to the structured textual modification of a grammar to adjust, refine, or improve its concrete syntactic form. In the literature, this process is also known as \emph{grammar transformation}~\cite{zhang2024supporting} or \emph{grammar optimization}~\cite{zhang2023automated}.
Grammar adaptation operates at multiple syntactic levels, including entire grammar rules, attributes, keywords, symbols (e.g., braces or parentheses), multiplicities, optionality, and other concrete elements.—and consists of adding, removing, or changing these textual elements to obtain a desired concrete syntax style or behavior~\cite{zhang2024supporting}. For example, Listing~\ref{lst:example} displays a grammar rule \texttt{Mission} containing an attribute \texttt{ownedComment}, whose keyword is identical to the attribute name. Through adaptation, the keyword in the attribute is removed (see Listing~\ref{lst:example_1}). 
% While grammar adaptation primarily concerns concrete-syntax refinements, it may intentionally introduce minor changes to the abstract syntax when the adaptation involves modifying assignments, types, multiplicities, or rule structure. 
\updated{While grammar adaptation primarily concerns concrete-syntax refinements, it may intentionally alter the abstract syntax tree structure derived from the grammar when the adaptation involves modifying assignments, types, multiplicities, or rule structure.}
In Xtext, such structural elements—rather than keywords or other concrete tokens—determine the resulting EMF-based abstract syntax tree~\cite{bettini2016implementing}.

\subsection{Rule-based Grammar Adaptation in the Context of Language Evolution}
% \todo[inline]{Get background knowledge from the previous work~\cite{zhang2024supporting,zhang2023automated}} %TODO BOWEN REVIEW on Monday, other sections review other days
The \textbf{\textit{rule-based}} grammar adaptation approach automatically transforms grammars generated from metamodels into target grammars that conform to specific styles and requirements through predefined transformation rules and configuration mechanisms. The core of this approach consists of two tools: \textit{GrammarTransformer}~\cite{zhang2024supporting} and \textit{ConfigGenerator}~\cite{zhang2023automated}.

\textit{GrammarTransformer} contains 60 predefined grammar transformation rules that operate on different elements of the grammar, including keywords, braces, optionality, multiplicity, grammar rule calls, etc. The transformation rules are divided into three types of operations: add, remove, and change. Each rule is associated with a well-defined scope that can be applied at different levels of granularity, such as the entire grammar, a specific grammar rule, or a specific attribute. Applying \textit{GrammarTransformer} requires language engineers to manually configure the application of transformation rules. Taking the transformation from Listing~\ref{lst:example} to Listing~\ref{lst:example_1} as an example, the following rules need to be configured: %need noun from 1-4
(1) \texttt{removeBraces} removes all curly braces in the $Mission$ rule; (2) \texttt{removeKeyword} removes all keywords such as $`Mission'$, $`ownedComment'$, etc.; (3) \texttt{removeOptionality} removes the optionality markers of attributes; (4) \texttt{removeComma} removes the comma in the attribute $ownedComment$; and etc. 
% These configurations are written as Java method calls, with each method accepting parameters to specify the target grammar rule, attribute name, etc.

\textit{ConfigGenerator} aims to automate the configuration of grammar transformation rules. It automatically extracts the required transformation rule configurations by comparing the generated grammar $G_1$ with the target grammar $G'_1$. 
% Its workflow consists of three steps: first, establishing a mapping relationship between grammar rules in $G_1$ and $G'_1$, then performing line-by-line comparisons for each matched pair of grammar rules, and finally identifying differences and selecting corresponding transformation rules. 
It automatically extracts transformation rule configurations by comparing $G_1$ with $G'_1$, which can then be reused for $G_2$.
The extracted configurations can be persisted and reused for the newly generated grammar $G_2$ in subsequent evolution steps, thereby automatically generating $G'_2$ (as shown in Figure~\ref{fig:problem}(b)). This approach achieved fully automated adaptation for languages such as EAST-ADL, BibTeX, and Xenia, but some grammar rules in DOT, Xcore, and SML still could not be fully automatically adapted. 
% mainly because certain complex grammar structures (such as syntactic predicates, nested optionality, order-insensitive attribute combinations, etc.) exceeded the coverage of predefined rules.

\section{Approach}
\label{sec:approach}
This section describes our LLM-based approach for automating grammar adaptation during metamodel evolution.

% We evaluate our approach on the same two DSLs used in prior work~\cite{zhang2024supporting}: QVTo, a model transformation language with four evolution versions, and EAST-ADL, an automotive domain language evaluated in both a simplified and a complete version. This choice enables direct comparison with the existing GrammarTransformer approach [1], a rule-based method, and provides realistic test cases for metamodel-grammar co-evolution.

\subsection{Overview}
Our approach addresses the regeneration scenario illustrated in Figure 1(b). It operates on four grammars: (i) $G_1$, the grammar generated from the original metamodel; (ii) $G'_1$, the grammar adapted from $G_1$; (iii) $G_2$, the grammar generated from the evolved metamodel; and (iv) $G'_2$, the target adapted grammar. The approach takes $G_1$, $G'_1$, and $G_2$ as inputs and outputs $G'_2$.

The approach proceeds in two steps. First, the LLM compares and analyzes $G_1$ and $G'_1$ to identify and learn the adaptations applied from $G_1$ to $G'_1$. Then, the LLM reuses these learned adaptations on $G_2$ to obtain $G'_2$. The expected $G'_2$ is intended to continue conforming to the evolved metamodel. We treat the internal processing of the LLM as a black box and evaluate only whether the resulting $G'_2$ conforms to the evolved metamodel and whether the adaptations from $G_2$ to $G'_2$ are consistent with or similar to those from $G_1$ to $G'_1$.

% Figure~\ref{fig:core_mechanism} illustrates the design rationale of our approach which aims to enable the LLM to automatically identify the adaptations reflected in the previous version of the grammar and to reuse these \updated{adaptations} in the grammar generated from the evolved metamodel (i.e., $G_2$), while ensuring that the adapted grammar (i.e., $G_2'$) remains consistent with the evolved metamodel. The design rationale can be summarized in two main points: 1) the LLM compares the original generated grammar (i.e., $G_1$) with its manually adapted version (i.e., $G_1'$) and learns the adaptations required to transform $G_1$ into $G_1'$; 2) when generate a new grammar from the evolved metamodel ($G_2$), the LLM model applies the learned adaptations to $G_2$ to generate the new adapted grammar $G_2'$ which is still consistent with the evolved metamodel.

\begin{figure}[t]
  \centering
  \includegraphics[width=0.7\linewidth]{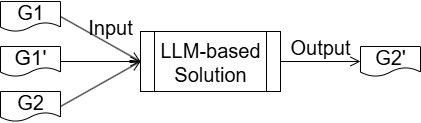}
  \caption{LLM-based grammar adaptation process}
  \label{fig:core_mechanism}
\end{figure}

\subsection{Experimental Setup and Prompting Strategy}
\label{sec:experi_set}
We evaluated our approach using three state-of-the-art LLMs: Claude Sonnet 4.5, Gemini 3, and ChatGPT 5.1, and conducted the experiments on six DSLs adopted from prior 
work~\cite{zhang2023automated}: EAST-ADL, BibTeX, Xenia, DOT, Xcore, and SML. Each 
included a grammar generated from its metamodel and a target grammar, representing diverse language characteristics, e.g., sizes, complexities, and domains. To develop a generalizable 
prompting strategy while mitigating the risk of overfitting, we split these 
into a \textit{training set} (EAST-ADL, BibTeX, Xenia, SML) and a \textit{test 
set} (DOT, Xcore). The training set was designed to cover diverse scenarios: 
small-scale (BibTeX with 43 grammar rules, Xenia with 15 rules), medium-scale (SML with 96 rules), and large-scale (EAST-ADL with 297 rules).

% \updated{On the training set, we iteratively refined the prompting strategy, experimenting with 
% different approaches to guide the LLM through the adaptation task.} 
For prompt design, we followed a manual, expert-in-the-loop \revised{development} protocol in which prompts were 
% iteratively revised based on observed model errors 
\revised{designed and validated}
on the training set, then frozen before final evaluation. The decision to \revised{accept or} revise a prompt was guided by evaluating the model's output quality based on the metamodel conformance and the adaptation consistency using metrics defined in Section~\ref{sec:eval_metrics}. This follows the human-guided refinement process described by Shah~\cite{shah2024prompt} and rooted in ``prompt programming'' principles (i.e., stepwise manual prompt improvement) introduced by Reynolds et al.~\cite{reynolds2021prompt}.

A single reference LLM, Claude Sonnet 4.5, was selected for prompt \revised{development} due to its demonstrated reliability in code-related tasks and strong performance in prior benchmarks~\cite{claude2025benchmark}. 
% Only this model was used during \revised{prompt development.}
% iterative improvement.
The core design remained consistent: the LLM received three grammar 
files ($G_1$, $G_1'$, $G_2$), compared $G_1$ and $G_1'$ to identify adaptations, 
and applied similar adaptations to $G_2$ to produce $G_2'$. 
\revised{We developed prompts starting from the small-to-medium scale DSLs (BibTeX, Xenia, SML), then validated their applicability on the large-scale case (EAST-ADL). The prompts focused on clearly communicating the two-stage task structure—first identifying adaptations from $G_1$ to $G_1'$, then applying them to $G_2$—
while relying on the prior-version grammar pair to guide the adaptation, rather than encoding task-specific heuristics in the prompt itself.}
% while relying on the provided grammar examples as the primary source of adaptation knowledge rather than encoding task-specific heuristics in the prompt itself.}
% We explored 
% variations in prompt formulation, including alternative strategies for providing the grammar inputs (sequential uploads 
% vs. simultaneous) and different phrasings of the adaptation request.
% We prioritized developing prompts on the small-to-medium scale DSLs (BibTeX, Xenia, SML) first, then validated their applicability on the large-scale 
% % EAST-ADL. During prompt development, if persistent difficulties arose with a particular LLM on EAST-ADL, we explored whether alternative LLMs in our evaluation set could better support the prompt refinement process.
% % For this exploratory phase, we used the web-based interfaces of the three LLMs.
% cases. 
For this exploratory phase, we used the web-based interface of Claude Sonnet 4.5.
\revised{For the cross-model evaluation, all three LLMs were evaluated manually through their respective web-based interfaces, with no additional automation. 
We focused on this setup rather than few-shot prompting, as example-based prompting can steer outputs toward the specific examples provided; by keeping the prompt decoupled from any particular DSL examples, we aimed to strengthen external validity and reduce overfitting to specific language characteristics.
Further details on the experiment execution procedure are available in our replication package~\cite{replication2025}.}
% \revised{The complete prompt development process, including all iterations and intermediate results, is documented in the replication package~\cite{replication2025}.}

\revised{Importantly, each DSL was processed in an independent LLM conversation session. Within each session, the LLM received only the three grammar files ($G_1$, $G_1'$, $G_2$) of that specific DSL; no information from other DSLs was carried over between sessions. The ``training set'' designation therefore refers to the DSLs used by the researchers to develop and validate the prompting strategy, not to any form of cross-DSL learning by the LLM. The training-test split follows standard evaluation practice to mitigate the risk of overfitting: prompts developed on the training DSLs might only work well on those specific languages, and the test set verifies that the finalized prompts generalize to unseen cases.}
% The finalized prompts were then evaluated across all three LLMs (Claude Sonnet 4.5, Gemini 3, and ChatGPT 5.1) to assess cross-model robustness and consistency, in line with recommendations from prompt sensitivity studies such as Zhuo et al.~\cite{zhuo2024prosa}.
The finalized prompts were evaluated across all three LLMs to assess cross-model robustness (Zhuo et al.~\cite{zhuo2024prosa}).

On the test set, we applied the finalized prompts without 
modification. Each generated grammar $G'_2$ was evaluated along two dimensions: metamodel conformance (i.e., whether $G_2'$ conformed to the evolved metamodel) and adaptation consistency (i.e., whether the adaptations from $G_1$ to $G_1'$ were performed on $G_2$). If syntactic or structural issues arose, we provided up to three targeted follow-up prompts (e.g., ``Add semicolons after 
attributes'')
% . This addressed: RQ1 - Can generalized prompts work across DSLs? (initial results); RQ2 - Can targeted refinement resolve issues? (refined results).
to assess whether targeted refinement could resolve the problems.
% \updated{Even if no issues arise, we still repeat the experiment three times for each LLM-based method on the test set DSLs to verify the stability of the results.}

Beyond the cross-sectional evaluation on test set DSLs, we conducted a longitudinal case study on QVTo to provide a more rigorous assessment of the LLM-based approach for supporting language evolution over multiple versions. QVTo offers a particularly relevant opportunity as it represents a real-world DSL with four official versions (1.0 through 1.3) documented by OMG~\cite{qvt2016omg}. This enables us to evaluate whether adaptations that LLM-based approaches learned from one version can be successfully reused across subsequent evolution transitions—a critical aspect of practical language evolution support. Specifically, we applied the finalized prompting strategy to each evolution transition (1.0 to 1.1, 1.1 to 1.2, 1.2 to 1.3) and compared the results against rule-based baselines from prior work, \revised{using Claude Sonnet 4.5 as the reference model.}

\subsection{Evaluation Metrics for Grammar Adaptation Quality}
\label{sec:eval_metrics}
% To systematically evaluate whether the LLM-generated adapted grammar $G_2'$ follows the same adaptations as $G_1'$, we designed a comprehensive set of evaluation metrics that assess both the technical correctness and adaptation consistency of the adapted grammars.
To evaluate whether $G_2'$ reproduces the adaptations from $G_1$ to 
$G_1'$, we define the following evaluation metrics.

\subsubsection{\revised{Grammar Rule-level} Adaptation Consistency (RAC)} 
RAC quantifies the proportion of grammar rules in $G_2'$ that successfully reproduce the adaptations observed in $G_1'$, 
calculated as RAC=$N_{\mathrm{correct}}$/$N_{\mathrm{total}}$.
% calculated as the Equation~\ref{eq:rac}. 
% \begin{equation}\label{eq:rac}
% \mathrm{RAC} = \frac{N_{\mathrm{correct}}}{N_{\mathrm{total}}}
% \end{equation}
$N_{\mathrm{correct}}$ denotes the count of grammar rules that were correctly adapted according to the previously learned adaptations, and $N_{\mathrm{total}}$ represents the total count of grammar rules that required adaptation.

\subsubsection{\revised{Grammar Rule-level} Output Similarity}
To quantify how closely adapted grammars match the target grammar, we adopt the evaluation framework from prior work~\cite{zhang2023automated}, i.e., for each DSL, we perform a rule-by-rule comparison between the adapted grammar and the target grammar, reporting the following measures: \textit{Same}: Number of grammar rules that are identical between the adapted grammar and the target grammar; \textit{Diff}: Number of grammar rules that differ between the two grammars; \textit{Percent}: Percentage of identical grammar rules (Same / Total × 100\%).
% \begin{itemize}
%     \item \textit{Same}: Number of grammar rules that are identical between the adapted grammar and the target grammar.
%     \item \textit{Diff}: Number of grammar rules that differ between the two grammars.
%     \item \textit{Percent}: Percentage of identical grammar rules (Same / Total × 100\%).
% \end{itemize}

For the LLM-based approach, we present both initial results (generalized prompts only) and refined results (after targeted follow-ups when needed). For the rule-based approach, we present the original results from~\cite{zhang2023automated}.
% \subsubsection{Comparison with Rule-based Approach from Previous Work}
% To further evaluate the applicability and potential of the proposed LLM-based approach, we applied it to the six case languages (EAST-ADL, BibTeX, Xenia, DOT, Xcore, and SML) used in the previous work~\cite{zhang2023automated}. The experiment followed the same input data and comparison procedure as the previous work, differing only in the grammar adaptation mechanism: while the previous work adapted grammars by extracting and replaying grammar optimization rule configurations, our approach enables the LLM to directly learn adaptations from the differences between the generated grammar and the target grammar, and then generate a new adapted grammar accordingly.

% During comparison, we reused the counting scheme from the previous work to perform grammar rule–level alignment between the adapted grammar and the target grammar, and we counted the number of grammar rules that were completely identical (Same), different (Diff), and the percentage of identical grammar rules among all grammar rules. This design allows for a direct comparison between the rule-based and LLM-based approaches without altering the original evaluation criteria.

% The goal of this comparative design is to observe the performance differences between the two approaches under identical data, alignment, and counting conditions, thereby providing an objective basis for assessing whether LLMs can serve as an alternative or complementary solution to rule extraction–based grammar adaptation.

\subsubsection{Metamodel Conformance}
In addition to adaptation consistency, we verify whether each adapted grammar $G_2'$ 
conforms to the metamodel $M_2$. This validation is necessary for the LLM-based approach, as the black-box nature of LLMs may 
produce grammars that are syntactically valid but semantically incompatible 
with the metamodel.

We validate metamodel conformance by: 1) loading $G_2'$ in the Eclipse Modeling Framework with Xtext; 2) checking for metamodel conformance violations; and 3) recording any violations such as undefined primitive types, missing or renamed attributes, etc.
% \begin{itemize}
%     \item Loading $G_2'$ in the Eclipse Modeling Framework with Xtext
%     \item Checking for metamodel conformance violations
%     \item Recording any violations such as undefined primitive types, missing or renamed attributes, and type mismatches in cross-references, etc.
% \end{itemize}
We report conformance results as binary (YES/NO) for each DSL, with detailed violation descriptions for non-conforming cases.

\section{Evaluation}
\label{sec:evaluation}
This section presents the experimental evaluation of our approach, covering prompt finalization, results on test set DSLs, longitudinal evolution support, and a comparative analysis of LLM-based and rule-based methods.

\begin{table*}[tb]
% \footnotesize
\scriptsize
\centering
\caption{Grammar Rule-Level Adaptation Correctness: LLM-based vs. Rule-based Approaches on Test Set DSLs}
\label{tab:correctness}
\begin{threeparttable}
\begin{tabular}{@{}lccccccccl@{}}
\toprule
\multirow{2}{*}{DSL} & \multirow{2}{*}{\begin{tabular}[c]{@{}l@{}}required\\ adaptations\tnote{1}\end{tabular}} & \multicolumn{2}{c}{Claude-Sonnet-4.5-Based Approach} & \multicolumn{2}{c}{ChatGPT-5.1-based Approach} & \multicolumn{2}{c}{Gemini-3-based Approach} & \multicolumn{2}{c}{Rule-based Approach} \\
\cmidrule(lr){3-4} \cmidrule(lr){5-6} \cmidrule(lr){7-8} \cmidrule(lr){9-10}
 &  & correct adaptations\tnote{2} & RAC & correct adaptations & RAC & correct adaptations & RAC & correct adaptations & RAC \\
\midrule
EAST-ADL & 234 & 37 & 15.81\% &  1 & 0.43\% & 139 & 59.4\% & 234 & 100\%\\
Xenia    & 15  & 15 & 100\%   & 15 & 100\%  & 15  & 100\%  & 43  & 100\%  \\
SML      & 96  & 96 & 100\%   & 96 & 100\%  & 96  & 100\%  & 94  & 97.91\% \\
\revised{BibTeX}   & 43  & 43 & 100\%   & 43 & 100\%  & 43  & 100\%  & 43  & 100\% \\
\midrule
\revised{DOT}      & 19  & 19 & 100\%   & 19 & 100\%  & 19  & 100\%  & 16  & 84.21\% \\
Xcore    & 32  & 32 & 100\%   & 32 & 100\%  & 32  & 100\%  & 20  & 62.50\% \\
\bottomrule
\end{tabular}
\begin{tablenotes}
\footnotesize
\item[1] The metric ``required adaptations'' refers to the count of grammar rules that need to be adapted from the generated grammar to the target grammar.
\item[2] The metric ``correct adaptations'' refers to the count of grammar rules that the approach correctly adapted.
\end{tablenotes}
\end{threeparttable}
\end{table*}

\subsection{Prompt Finalization and Experimental Environment}
\label{sec:prompt_dev}
% We conducted experiments on \warning{to be completed} using Claude 4 Sonnet and GPT-4o. 
% The evaluation in this study involved a total of six case languages. Among them, the authors directly adopted the six case languages from prior work~\cite{zhang2023automated}, namely EAST-ADL, BibTeX, Xenia, DOT, Xcore, and SML. Each of them contains a grammar generated from the metamodel and a target grammar. These six DSLs were used directly according to the original datasets and comparison procedures, without constructing additional versions.
The decision to finalize a prompt was strictly governed by meeting a set of objective quality thresholds on the training set. For a prompt to be accepted and frozen for final evaluation, the LLM output on the training set had to satisfy three criteria: the adapted target grammar $G_2'$ must conform to the metamodel $M_2$ (i.e., the grammar can be fully resolved), and both the RAC value and the \revised{grammar rule-level} output similarity (Percent) value exceed $90\%$.

% All experiments were conducted using Claude Sonnet 4.5. 
Metamodel conformance of the generated adapted grammars was validated employing the Eclipse Modeling Tools without enforcing strict version requirements. The authors used the March 2024 release, Java 17, and Xtext 2.36.0. 
% The experimental scripts were implemented in Python 3.11. The Python scripts implementing our approach and 
All experimental data are available in our replication package at~\cite{replication2025}.

\subsection{Grammar Adaptation Results (RQ1)}
\label{sec:eval_results}

% Through iterative refinement on the training set, we finalized two prompts for our approach. 
\revised{On BibTeX, Xenia, and SML, general-purpose two-step prompts met all three quality thresholds without requiring refinement. We then applied the same prompts to the large-scale case (i.e., EAST-ADL with 297 grammar rules) and conducted five iterations with progressively more specific instructions; however, RAC remained far below 90\% (see the analysis in Section~\ref{sec:limit_llm}). Based on these results, we took the general-purpose two-step prompts that had met all quality thresholds on the three small-to-medium scale DSLs as the finalized prompts. 
% As shown below, the finalized prompts reflect the design rationale described in Section~\ref{sec:experi_set}.
}
Prompt 1, used when providing $G_1$ and $G_1'$, and instructs the LLM to identify and extract adaptation rules by comparing the two grammars.  
Prompt 2, used when providing $G_2$, instructs the LLM to apply the previously learned adaptations to produce the adapted grammar $G_2'$. 
% These prompts were developed through systematic experimentation on BibTeX, Xenia, and SML. 
% \updated{These prompts were developed through systematic experimentation on BibTeX, Xenia, and SML. 
% While EAST-ADL was initially included in prompt development, results remained below 90\% RAC despite extensive iterative refinement attempts. We therefore excluded it from finalization, achieving RAC > 90\%, output similarity > 90\%, and metamodel conformance across the three smaller DSLs before freezing the prompts for evaluation. 
For completeness and transparency, we also applied the finalized prompts to EAST-ADL as a large-scale case, following the same evaluation procedure used for the test set DSLs (i.e., allowing up to three follow-up prompts).
% we report the fifth iteration results for EAST-ADL 
The results reported
in Tables~\ref{tab:correctness}-\ref{tab:conformance}
% , which represent our best-effort attempt before determining that  
for EAST-ADL are drawn from this evaluation, and reflect
\revised{the scalability limitations of current LLMs, rather than prompt design, that prevented achieving acceptable quality at}
% approach was not suitable for 
this scale of grammar~\revised{(see Section~\ref{sec:limit_llm}).}
% EAST-ADL was excluded from finalization after multiple iterative attempts failed to achieve consistent quality thresholds due to its scale and complexity - 
% achieving RAC > 90\%, output similarity > 90\%, and metamodel conformance across all three DSLs before being frozen for evaluation.

% \begin{tcolorbox}[colback=white!100!black, colframe=white!50!black, arc=0mm, left=0.5em, right=0.5em, top=0.5em, bottom=0.5em, label=prompt:identify]
% % \small
% \updated{\textbf{Prompt 1}: The attachment contains two Xtext grammars for the same language: the grammar generated from the meta-model and the target grammar. Please identify the adaptations required to transform the generated grammar into the target grammar.}
% \end{tcolorbox}

% \begin{rqbox}{Prompt 1}
% The attachment contains two Xtext grammars for the same language: the grammar generated from the meta-model and the target grammar. Please identify the adaptations required to transform the generated grammar into the target grammar.
% \end{rqbox}
\begin{promptbox}
\textbf{Prompt 1:} The attachment contains two Xtext grammars for the same language: the grammar generated from the metamodel and the target grammar. Please identify the adaptations required to transform the generated grammar into the target grammar.
\end{promptbox}

% \begin{tcolorbox}[colback=white!100!black, colframe=white!50!black, arc=0mm, left=0.5em, right=0.5em, top=0.5em, bottom=0.5em, label=prompt:apply]
% % \small
% \updated{\textbf{Prompt 2}: Now, I'm sending you the grammar generated from the evolved metamodel. Please adapt it using the adaptations you learned previously and output the adapted grammar to me.}
% \end{tcolorbox}

% \begin{rqbox}{Prompt 2}
% Now, I'm sending you the grammar generated from the evolved metamodel. Please adapt it using the adaptations you learned previously and output the adapted grammar to me.
% \end{rqbox}

\begin{promptbox}
\textbf{Prompt 2:} Now, I'm sending you the grammar generated from the evolved metamodel. Please adapt it using the adaptations you learned previously and output the adapted grammar to me.
\end{promptbox}

\begin{table*}[tb]
% \footnotesize
\scriptsize
\centering
\begin{threeparttable}
\caption{Grammar Rule-level Similarity Comparison between LLM-based/Rule-based Adapted Grammars and Target Grammars on Test Set DSLs}
\label{tab:similarity}
\begin{tabular}{@{}lcccccccccccl@{}}
\toprule
\multirow{2}{*}{DSL} & \multicolumn{3}{c}{Claude-Sonnet-4.5-based Approach} & \multicolumn{3}{c}{ChatGPT-5.1-based Approach} & \multicolumn{3}{c}{Gemini-3-based Approach} & \multicolumn{3}{c}{Rule-based Approach} \\
\cmidrule(lr){2-4} \cmidrule(lr){5-7} \cmidrule(lr){8-10} \cmidrule(lr){11-13}
 & Same\tnote{1} & Diff & Percent\tnote{2} & Same & Diff & Percent & Same & Diff & Percent & Same & Diff & Percent \\
\midrule
EAST-ADL & 100 & 197 & 33.67\% &  1 & 296 & 0.34\% & 202 & 95 & 68.01\% & 297 & 0  & 100\%\\
Xenia    & 15  & 0   & 100\%   & 15 & 0   & 100\%  & 15  & 0  & 100\%   & 15  & 0  & 100\% \\
SML      & 75  & 0   & 100\%   & 75 & 0   & 100\%  & 75  & 0  & 100\%   & 51  & 24 & 68.0\%\\
BibTeX   & 43  & 0   & 100\%   & 43 & 0   & 100\%  & 43  & 0  & 100\%   & 43  & 0  & 100\% \\
\midrule
\revised{DOT}      & 24  & 0   & 100\%   & 24 & 0   & 100\%  & 24  & 0  & 100\%   & 21  & 3  & 87.50\% \\
Xcore    & 40  & 0   & 100\%   & 40 & 0   & 100\%  & 40  & 0  & 100\%   & 28  & 12 & 70.00\% \\
\bottomrule
\end{tabular}
\begin{tablenotes}
\item[1] The metric ``Same'' refers to the count of grammar rules in the adapted grammar that are identical to those in the target grammar. The metric ``Diff'' is the same principle.
\item[2] The metric ``Percent'' refers to the proportion of the count of identical grammar rules in the adapted grammar and the target grammar to the total count of grammar rules in the target grammar.
\end{tablenotes}
\end{threeparttable}
\end{table*}

To address \textbf{RQ1}, \updated{Tables~\ref{tab:correctness}-\ref{tab:conformance} present evaluation results across all six DSLs in terms of RAC, output similarity, and metamodel conformance. We focus our analysis primarily on the test set DSLs (DOT and Xcore), as these results—obtained by applying finalized prompts to unseen languages—demonstrate generalization capability. Training set results are included in Tables 1-3 for completeness but are not the focus of analysis, as prompt development inherently used these DSLs.}

Table~\ref{tab:correctness} presents a comparison of \revised{grammar rule-level} adaptation correctness between three LLM-based methods and the rule-based method on the test set DSLs (DOT and Xcore). In DOT, 19 grammar rules need to be adapted from the generated grammar to the target grammar, while in Xcore, 32 grammar rules require adaptation. From the RAC metric perspective, all three LLM-based methods achieved 100\% adaptation correctness on both test set DSLs. In contrast, the rule-based method achieved an RAC value of 84.21\% on DOT (16 out of 19 rules correctly adapted) and only 62.50\% on Xcore (20 out of 32 rules correctly adapted). These results indicate that on the test set DSLs, the LLM-based methods demonstrate superior performance in identifying and applying adaptations learned from the training set compared to the rule-based method.
\updated{On the training set, LLM-based approaches achieved high RAC (100\%) on BibTeX, Xenia, and SML, while the rule-based approach achieved 100\% on EAST-ADL, Xenia, and BibTeX but 97.91\% on SML. EAST-ADL posed challenges for LLM-based approaches (see Section~\ref{sec:limit_llm} for in-depth analysis).}

Table~\ref{tab:similarity} further presents the similarity comparison results between the adapted grammars and the target grammars. The results show that all three LLM-based methods achieved 100\% similarity on both test set DSLs, meaning that the grammars adapted by the LLM-based methods are identical to the target grammars in all grammar rules. This result perfectly echoes the RAC values in Table 1. In contrast, the rule-based method achieved 87.50\% similarity on DOT, with 21 identical rules and 3 differing rules compared to the target grammar, and only 70.00\% on Xcore (28 identical rules and 12 differing rules). These differences primarily stem from the rule-based method's inability to handle certain complex adaptations, such as the predicated assignment structures in DOT and the order-insensitive boolean attribute combinations in Xcore (see Section~\ref{sec:comp_ana} for details).
\updated{On the training set, LLM-based approaches achieved 100\% similarity on BibTeX, Xenia, and SML, while still, faced challenges in EAST-ADL. The rule-based approach achieved 100\% on EAST-ADL, Xenia, and BibTeX but only 68.0\% on SML.}

\updated{Table~\ref{tab:conformance} presents the metamodel conformance validation results of the adapted grammars. On the test set, all three LLM-based methods passed metamodel conformance validation on both DOT and Xcore, while the rule-based method failed validation on Xcore. On the training set, LLM-based methods passed validation on BibTeX, Xenia, and SML, but encountered challenges on EAST-ADL; the rule-based method failed validation on SML, though it succeeded on EAST-ADL and BibTeX. These results indicate that on the test set DSLs, the adapted grammars generated by the LLM-based methods are not only consistent with the target grammars at the textual level, but also satisfy all constraints of the metamodel at the semantic level, demonstrating practical usability. However, the failure of LLM-based methods on EAST-ADL suggests potential limitations when handling large-scale grammars (see Section~\ref{sec:limit_llm} for in-depth analysis).}

To verify the stability of the results, we repeated the experiments three times for each LLM-based method on the test set DSLs, and all runs yielded consistent results.

% \begin{answerbox}{Answer to RQ1}
% On the test set of small-to-medium scale DSLs, all three LLMs achieved 100\% \revised{grammar rule-level} adaptation consistency and output similarity, demonstrating that LLMs can learn and apply grammar adaptations from historical versions; however, on large grammars (EAST-ADL, 297 rules), adaptation consistency was far below 90\%, revealing the limitations of current LLMs in scaling.
% \end{answerbox}

\begin{answerbox}
\textbf{Answer to RQ1:} On the test set of small-to-medium scale DSLs, all three LLMs achieved 100\% RAC and output similarity, demonstrating that LLMs can learn and apply grammar adaptations from historical versions; however, on large grammars (EAST-ADL, 297 rules), adaptation consistency was far below 90\%, revealing the limitations of current LLMs in scaling.
\end{answerbox}

% The results show that while LLM-generated adapted grammars were largely successful, they were not error-free. For OpenAI, Claude-4 produced 2 errors and GPT-4o produced 0 errors in the adapted grammars. For majordomo, Claude-4 produced 3 errors and GPT-4o also produced 3 errors. However, the number of errors in most cases was relatively small and could be easily corrected manually, suggesting that LLM-based approaches show promise in reducing the manual effort required for grammar adaptation during metamodel evolution, though some post-processing may still be needed.

% In terms of pattern consistency metrics, Claude-4 achieved perfect rule-level adaptation consistency (RAC = 100\%) and complete structural transformation preservation (STP = 1) for OpenAir, though both models showed lower pattern generalization capability (PGS = 0.2) for this DSL. For majordomo, Claude-4 demonstrated 85\% RAC with strong pattern generalization (PGS = 1) and high structural transformation preservation (STP = 0.91). GPT-4o's performance on majordomo was notably weaker, with 60\% RAC, perfect STP (1.0), but no pattern generalization capability (PGS = 0). These variations suggest that the effectiveness of LLM-based grammar adaptation depends on both the specific LLM model and the characteristics of the DSL being adapted.
\subsection{Language Evolution Support: A Longitudinal Study on QVTo (RQ2)}
\label{sec:evolution}

To address \textbf{RQ2}, Table~\ref{tab:qvto_evolution} presents the comparative results of LLM-based and rule-based approaches across QVTo's evolution sequence.
\revised{This study was conducted using Claude Sonnet 4.5 (the reference model for prompt development), as its focus is cross-version reusability rather than cross-model robustness—the latter being evaluated in Section~\ref{sec:eval_results}.}
Both approaches achieved 100\% RAC across all three evolution steps, demonstrating their capability to successfully maintain grammar adaptations during language evolution driven by metamodel changes. The key distinction lies in required human effort: the LLM-based approach required no manual grammar editing across all evolution steps, whereas the rule-based approach necessitated configuration adjustments in two out of three transitions (two changes for V1.0$\rightarrow$V1.1 and one change for V1.2$\rightarrow$V1.3). This suggests that for DSLs of moderate scale like QVTo, LLM-based adaptation can potentially achieve high reusability across evolution steps once a suitable prompting strategy is established.
\revised{All adapted grammars passed metamodel conformance validation across all three evolution steps.}

% \begin{answerbox}{Answer to RQ2}
% Across three evolution steps in QVTo, the LLM-based approach 
% % achieved 100\% adaptation consistency and output similarity with 
% \revised{successfully reused learned adaptations across all consecutive evolution steps without manual grammar editing,}
% % zero human intervention, 
% demonstrating its capability to \revised{reuse adaptation strategies across multiple evolution steps.}
% % support multi-version language evolution 
% % with minimal human effort.
% \end{answerbox}

\begin{answerbox}
\textbf{Answer to RQ2:} Across three evolution steps in QVTo, the LLM-based approach 
% achieved 100\% adaptation consistency and output similarity with 
\revised{successfully reused learned adaptations across all consecutive evolution steps without manual grammar editing,}
% zero human intervention, 
demonstrating its capability to \revised{reuse adaptation strategies across multiple evolution steps.}
\end{answerbox}

\begin{table}[tb]
% \footnotesize
\scriptsize
\centering
\begin{threeparttable}
\caption{Metamodel Conformance Validation of Adapted Grammars on Test Set DSLs}
\label{tab:conformance}
\begin{tabular}{@{}lcccc@{}}
\toprule
DSL & Claude-based & ChatGPT-based & Gemini-based & Rule-based \\
\midrule
EAST-ADL & NO & NO & NO & YES \\
Xenia & YES & YES & YES & YES \\
SML & YES & YES & YES & NO \\
\revised{BibTeX} & YES & YES & YES & YES \\
\midrule
\revised{DOT} & YES & YES & YES & YES \\
Xcore & YES & YES & YES & NO \\
\bottomrule
\end{tabular}
% \begin{tablenotes}
% \item[a] Your note here
% \end{tablenotes}
\end{threeparttable}
\end{table}

\subsection{Comparative Analysis: Strengths and Limitations (RQ3)}
\label{sec:comp_ana}
To address \textbf{RQ3}, we perform a comparative analysis to identify in which adaptation scenarios LLM-based approaches outperform rule-based methods, and in which scenarios they encounter challenges.

\subsubsection{Advantages of the LLM-based Approach: Adapting Complex Grammar Scenarios}
\label{sec:adavantage_llm}
% As shown in Section~\ref{sec:eval_results}, the LLM-based approach outperforms the rule-based approach \updated{particularly} on the test set DSLs, achieving 100\% RAC and output similarity on both DOT and Xcore, while also passing metamodel conformance validation. In contrast, the rule-based approach achieved an RAC of 84.21\% on DOT and 62.50\% on Xcore, and failed metamodel conformance validation on Xcore. Furthermore, the longitudinal study in Section~\ref{sec:evolution} demonstrates that across three evolution steps in QVTo, the LLM-based approach achieved fully automated adaptation without manual grammar editing, while the rule-based approach required manual configuration adjustments in two of the three evolution steps. 
As shown in Sections ~\ref{sec:eval_results}-\ref{sec:evolution}, the LLM-based approach achieved 100\% RAC and output similarity on both test set DSLs while passing metamodel conformance validation, outperforming the rule-based approach (84.21\% on DOT, 62.50\% on Xcore, conformance failure on Xcore). In QVTo, it achieved fully automated adaptation across all three evolution steps, while the rule-based approach required manual adjustments in two.
% These results from both the test set and longitudinal study prompt us to analyze in depth: in these successful cases, what capabilities does the LLM-based approach possess that the rule-based approach lacks?

In-depth analysis of the failure cases of the rule-based approach on the test set reveals its fundamental limitation: it relies on predefined rules for matching based on surface features and cannot understand the contextual intent and semantic relationships of grammar structures. In DOT, the rule-based approach fails in three types of scenarios: when encountering the syntactic predicate \texttt{=>}, it cannot recognize its role in backtracking parsing — in the Port rule shown in Listing~\ref{lst:syntactic-predicate}, \texttt{=>} instructs the parser to perform backtracking among multiple choice branches, but the rule-based approach cannot understand how to adapt this special construct; when facing multiple optionality (such as the nested inner and outer optionality in \texttt{('subgraph' (name=ID)?)?)}, it cannot coordinate the hierarchical relationships; when handling multi-symbol ``or'' combinations, it cannot determine the overall optionality.

\begin{lstlisting}[basicstyle=\footnotesize\ttfamily, breaklines=true, label=lst:syntactic-predicate, caption={Port rule in DOT grammar containing syntactic predicate.}]
Port returns Port:
    {Port}
    ':'
    (=> compass_pt=COMPASS_PT |
        name=ID |
        name=ID ":" compass_pt=COMPASS_PT);
\end{lstlisting}

\begin{table*}[tb]
\centering
\footnotesize
\caption{Evolution Step Analysis on QVTo \revised{(LLM-based: Claude Sonnet 4.5)}}
\label{tab:qvto_evolution}
\begin{threeparttable}
\begin{tabular}{@{}llllllll@{}}
\toprule
\multicolumn{4}{c}{QVTo Evolution Context} & \multicolumn{2}{c}{\revised{Claude}-based Approach} & \multicolumn{2}{c}{Rule-based Approach} \\
\cmidrule(lr){1-4} \cmidrule(lr){5-6} \cmidrule(lr){7-8}
Versions & Grammar Rules & Evolution Step & Metamodel Changes & RAC & Follow-ups & RAC & \#cORA \\
\midrule
V1.0 & 1026 & - & - & - & - & - & - \\
V1.1 & 992 & V1.0 $\rightarrow$ V1.1 & 29 differences & 100\% & 0 & 100\% & 2 \\
V1.2 & 992 & V1.1 $\rightarrow$ V1.2 & 0 differences (QVTo part)\tnote{a} & 100\% & 0 & 100\% & 0 \\
V1.3 & 991 & V1.2 $\rightarrow$ V1.3 & 1 difference & 100\% & 0 & 100\% & 1 \\
\bottomrule
\end{tabular}
\begin{tablenotes}
\small
\item[a] Version 1.2 introduced three metamodel changes, but all were in the excluded OCL part.
\end{tablenotes}
\end{threeparttable}
\end{table*}

In Xcore, the rule-based approach similarly fails in four types of situations: it cannot recognize the structural intent of mutually exclusive alternatives, cannot handle attribute movement across same-named keywords, and although it can identify individual operations in a combination, it cannot determine the correct execution order—as shown in Listing~\ref{lst:correct-order}, the bounds attribute requires multiple operations such as removing curly braces, renaming keywords, and changing delimiters; while the rule-based approach can identify these independent operations, it cannot determine their correct execution order, leading to adaptation failure. Additionally, when cross-reference types are missing, the rule-based approach is also disrupted and unable to adapt—in Listing~\ref{lst:cross-reference}, 
the type attribute in the generated grammar is missing one cross-reference type, which prevents the rule-based 
% approach from identifying the complete adaptation rules required to transform the generated grammar to the target grammar.
approach's adaptation.

\begin{lstlisting}[basicstyle=\footnotesize\ttfamily, breaklines=true, label=lst:correct-order, caption={Example of combination operation ordering failure in Xcore.}]
// Generated grammar
XTypeParameter returns XTypeParameter:
	{XTypeParameter}
	'XTypeParameter'
	name=EString
	'{'
		('annotations' '{' annotations+=XAnnotation ( "," annotations+=XAnnotation)* '}' )?
		('bounds' '{' bounds+=XGenericType ( "," bounds+=XGenericType)* '}' )?
	'}';

// Target grammar
XTypeParameter returns XTypeParameter:
    {XTypeParameter}
          (annotations+=XAnnotation)*  
    name=ID
        ('extends'  bounds+=XGenericType ( "&" bounds+=XGenericType)*  )?
    ;
\end{lstlisting}

\begin{lstlisting}[basicstyle=\footnotesize\ttfamily, breaklines=true, label=lst:cross-reference, caption={Example of adaptation failure due to missing cross-reference type in Xcore.}]
// Generated grammar - cross-reference type missing
XGenericType returns XGenericType:
    {XGenericType}
    'XGenericType'
    '{'
        ('type' type=[|EString])?
        ...
    '}';

// Target grammar - complete cross-reference type
XGenericType returns XGenericType:
    {XGenericType}
    type=[genmodel::GenBase|XQualifiedName]
    ...;
\end{lstlisting}

These failures reveal a common problem: correct adaptation requires not only identifying ``what grammar elements exist'' but also understanding ``what intent these elements express in this context''. For example, the symmetric choice structure in Xcore shown in Listing~\ref{lst:order-insensitive} is not a regular mutually exclusive choice, but rather expresses order insensitivity; multiple optionality in DOT is not simple stacking, but rather hierarchical structural relationships. The 60 predefined rules of the rule-based approach are based on surface features of grammar elements and cannot capture these deep design intents.

\begin{lstlisting}[basicstyle=\footnotesize\ttfamily, breaklines=true, label=lst:order-insensitive, caption={Example of order-insensitive boolean attribute combination.}]
    ...
    (unordered?='unordered' unique?='unique'?|
    unique?='unique' unordered?='unordered'?)?
    ...
\end{lstlisting}

In contrast, the LLM-based approach learns context-relevant handling strategies from examples by analyzing the complete adaptation process from $G_1$ to $G_1'$. LLMs do not need to explicitly understand the semantic definitions of syntactic predicates or multiple optionality, but rather observe ``when a certain structure appears in historical adaptations, how it was handled'', thereby implicitly learning strategies such as preserving special constructs, coordinating hierarchical relationships, and maintaining structural intent. This ability to learn from complete adaptation examples not only enables LLMs to handle complex scenarios that the predefined rules of the rule-based approach cannot cover, but also demonstrates better generalization capability—as the QVTo longitudinal study shows, once an LLM learns adaptations from one evolution step, it can automatically reuse these adaptations in subsequent evolutions without manual grammar editing.

% Accurate Adaptation Recognition and Over-generalization Issues. Xenia primarily involves systematic, rule-based transformations (e.g., removing container braces, keyword standardization), which the rule-based approach naturally handles.
% % through its 60 predefined optimization rules. 
% However, detailed analysis reveals that the LLM performed many erroneous keyword renaming operations during the adaptation process—renamings that do not exist in the adaptations from G1 to G'1—leading to unnecessary deviations between the generated grammar and the target grammar, indicating that the LLM may over-generalize learned adaptations. SML performed poorly with both approaches (68\% and 12\%). The difficulty for the rule-based approach may stem from the language's use of embedded expressions~\cite{zhang2023automated}, while the LLM approach's low accuracy may be related to the substantial differences between the generated grammar and the target grammar—57 grammar rules were removed during the adaptation process, and such large-scale structural changes may exceed the LLM's ability to learn adaptations from examples.

Furthermore, the LLM-based approach is more reliable in ensuring metamodel conformance. Although the rule-based approach achieved 70\% textual similarity on Xcore, the generated grammar does not conform to the metamodel. This is because transformations based on surface patterns may break the mapping relationship between grammar and metamodel. In contrast, by learning the complete adaptation process $G_1$$\rightarrow$$G_1'$ that conforms to the metamodel, LLMs implicitly learn the constraints for maintaining conformance, resulting in grammars that are both textually correct and semantically compliant with the metamodel.

\subsubsection{Limitations of the LLM-based Approach: Challenges with Large-Scale Grammars}
\label{sec:limit_llm}
Although the LLM-based approach performed well on the test set DSLs, it encountered difficulties on the EAST-ADL language in the training set. The RAC values of the three LLM-based approaches on EAST-ADL were all far below 90\%, and acceptable adaptation quality could not be achieved even after multiple iterations of prompt optimization (as we mentioned in Section~\ref{sec:prompt_dev}, the results presented in Tables~\ref{tab:correctness}-\ref{tab:conformance} for EAST-ADL represent the fifth iteration of our prompt refinement). EAST-ADL is the largest among all evaluated DSLs, containing 291 metaclasses, 297 grammar rules, and approximately 3000 lines of grammar text. We found that grammar scale may be a key factor affecting the performance of the LLM-based approach: on large-scale grammars like EAST-ADL, LLMs omit a large number of identified adaptation operations during the application phase. 

In the third iteration, Claude successfully identified approximately 1500 semantic adaptations, including moving the shortName attribute from inside curly braces to outside, adding semicolons after attributes with multiplicity ``0-1'', removing comma separators from collection attributes, and other key adaptations. 
% However, 
\revised{This suggests that, for Claude, the primary bottleneck was not in the identification stage—where the LLM analyzes $G_1$ and $G_1'$ to recognize adaptation operations—but rather in the application stage, where identified adaptations must be consistently executed across all grammar rules. Indeed,}
when Claude was asked to apply these identified adaptations to the evolved grammar $G_2$, the generated adapted grammar systematically omitted a large number of identified adaptations, especially the addition of semicolons. 
% The reasons behind this phenomenon are worth exploring in depth. 
The three grammar files of EAST-ADL ($G_1$, $G_1'$, $G_2$) produced a total of approximately 12,000 tokens, which is far below Claude Sonnet 4.5's context window limit of 200,000 tokens, but when adaptations need to be applied to $G_2$ containing hundreds of grammar rules, LLMs seem to have difficulty maintaining consistency in detail conventions across the entire grammar scope. 

\begin{table}[tb]
\caption{Adaptation-type-level analysis on EAST-ADL: correctness of each adaptation type across LLMs.}
\label{tab:adaptation-type}
\resizebox{\columnwidth}{!}{%
\begin{threeparttable}
\begin{tabular}{lrrrrrrrr}
\toprule
 & & \multicolumn{2}{c}{\textbf{Claude Sonnet 4.5}} & \multicolumn{2}{c}{\textbf{Gemini 3}} & \multicolumn{2}{c}{\textbf{ChatGPT 5.1}} \\
\cmidrule(lr){3-4} \cmidrule(lr){5-6} \cmidrule(lr){7-8}
\textbf{Adaptation Type\tnote{a}} & \textbf{Occ.\tnote{b}} & \textbf{Cor.\tnote{c}} & \textbf{Inc.\tnote{d}} & \textbf{Cor.} & \textbf{Inc.} & \textbf{Cor.} & \textbf{Inc.} \\
\midrule
Brace/optionality removal   & 194 &  78 & 116 & 193 & 1  & 2 & 192 \\
    Keyword removal         & 194 & 190 & 4   & 191 & 3  & 3 & 191 \\
shortName promotion         & 188 & 188 & 0   & 134 & 54 & 2 & 186 \\
Separator modification      & 225 &  47 & 178 & 180 & 45 & 5 & 220 \\
Type system adaptation      & 202 & 151 & 51  & 201 & 1  & 5 & 197 \\
\bottomrule
\end{tabular}
\begin{tablenotes}
\footnotesize
\item[a] Adaptation types are derived from the grammar transformation rules identified in~\cite{zhang2024supporting}.
\item[b] Occ.: number of grammar rules requiring this adaptation type.
\item[c] Cor.: number of grammar rules where the adaptation was correctly applied.
\item[d] Inc.: number of grammar rules where the adaptation was not correctly applied.
\end{tablenotes}
\end{threeparttable}%
}
\end{table}

It is worth noting that this limitation is unrelated to grammar complexity—EAST-ADL's adaptations are relatively systematic (e.g., applying the same shortName promotion and curly brace optionalization to all entity rules), and should theoretically be easier to learn than the complex adaptations involving special constructs in DOT and Xcore. The real challenge lies in scale: 
\revised{the six evaluated DSLs exhibit an inverse relationship between grammar scale and adaptation complexity: DOT and Xcore are small-to-medium in scale but involve complex, context-dependent adaptations (syntactic predicates, predicated assignments, order-insensitive attribute combinations), whereas EAST-ADL is large in scale but requires repetitive, uniform adaptations across rules (Table~\ref{tab:adaptation-type}). This distinction helps explain why LLMs succeeded on the former but struggled with the latter—the challenge is scale-dependent rather than complexity-dependent.}

Gemini 3's failure on EAST-ADL exhibited different problem characteristics: in addition to the same systematic omission of adaptations, Gemini also performed erroneous destructive operations, including deleting the shortName attribute that should not be deleted, modifying attribute names without reason causing attributes to be unparseable, and adding semicolons where semicolons should not be added. These errors resulted in the generated grammar not only having an RAC value below 90\%, but also failing metamodel conformance validation.
% , making it completely unusable in practice. 
Multiple targeted follow-up prompts (such as explicitly requesting "do not delete attributes that should not be deleted") also failed to resolve these issues.
% Gemini 3's failure on EAST-ADL was primarily caused by its inability to output the complete adapted grammar in a single response. The output was split across multiple segments, and even after merging, a big number of grammar rules and lines within rules were missing. Additionally, Gemini reordered grammar rules unpredictably, increasing the difficulty of comparison. Follow-up instructions requesting complete output, preservation of all rules, and maintenance of rule ordering were consistently ignored. This is consistent with Table~\ref{tab:adaptation-type}, which shows zero correct adaptations across all five types for Gemini 3.

ChatGPT 5.1 \revised{encountered additional challenges on EAST-ADL: its input was truncated to approximately 1,200 of 3,000 lines, preventing it from receiving the complete grammar.}
% its maximum input per message is approximately 30K-32K tokens, while EAST-ADL's generated grammar of approximately 3000 lines could only receive about 1200 lines. 
This means that in the first step of identifying required adaptations, 
% ChatGPT had already missed most of the information, causing the subsequent adaptation application to fundamentally lose its foundation.
ChatGPT had already missed most of the grammar content, making accurate adaptation impossible.

\revised{To quantify these observations, we grouped the adaptation operations required to transform the generated grammar into the target grammar of EAST-ADL into five adaptation types based on their operation characteristics, and analyzed correctness by adaptation type (Table~\ref{tab:adaptation-type}). Results show that Claude's difficulties lie in consistently applying operations: keyword removal achieved high correctness (190 out of 194 occurrences), but separator modification succeeded in only 47 out of 225 cases. Gemini achieved high correctness on most types but struggled with shortName promotion (134 out of 188) and separator modification (180 out of 225), with some failures involving deletion of attributes that should have been preserved, causing metamodel conformance failures. ChatGPT's performance was limited by input truncation as discussed earlier. 
% These results suggest that the challenges lie in consistently applying adaptation operations across hundreds of grammar rules rather than in identifying individual operation types. 
These results confirm that the bottleneck lies in the application stage rather than in identification: Claude and Gemini were able to describe the required adaptation types in their intermediate outputs, yet failed to execute them completely across all grammar rules; ChatGPT, by contrast, could not complete the identification stage itself due to input truncation.
Despite refinement attempts through follow-up instructions, the systematic omission persisted, suggesting that this limitation reflects the capability boundaries of the LLMs evaluated in this study, rather than a prompt engineering issue that could be resolved through further refinement.}

% It is worth noting that t
The rule-based approach achieved 100\% adaptation correctness (RAC) on EAST-ADL. This is because once adaptation rules are determined, the rule-based approach can systematically apply the rules to all grammar rules in a programmatic manner, unaffected by grammar scale. This contrast indicates that when handling large-scale grammar adaptation tasks with high repetition of identical operations, the reliability and scalability of the rule-based approach are superior to the current LLM-based approach.

% \begin{answerbox}{Answer to RQ3}
% LLM-based approaches demonstrate advantages in handling complex grammar scenarios that are difficult for rule-based methods to address, such as syntactic predicates, predicated assignments, and order-insensitive attribute combinations, successfully learning and applying adaptations in these cases. However, LLM-based approaches encountered challenges on large-scale grammars, whereas rule-based approaches demonstrate better reliability and scalability in adapting large languages.
% \end{answerbox}

\begin{answerbox}
\textbf{Answer to RQ3:} LLM-based approaches demonstrate advantages in handling complex grammar scenarios that are difficult for rule-based methods to address, such as syntactic predicates, predicated assignments, and order-insensitive attribute combinations, successfully learning and applying adaptations in these cases. However, LLM-based approaches encountered challenges on large-scale grammars, whereas rule-based approaches demonstrate better reliability and scalability in adapting large languages.
\end{answerbox}

\section{Discussion}
\label{sec:discussion}
\subsection{\revised{Practical Guidelines}}
\revised{Based on our evaluation, we provide the following guidelines for selecting adaptation approaches:}

\revised{For grammars with fewer than 100 rules, LLM-based approaches are recommended when adaptations involve syntactic predicates, predicated assignments, or order-insensitive attribute combinations—constructs that require extensive manual configuration in rule-based methods. For grammars exceeding 200 rules, rule-based approaches are more reliable, as our EAST-ADL evaluation (297 rules, Table~\ref{tab:adaptation-type}) demonstrates LLMs' systematic omission issues at this scale.}

\revised{The choice between approaches should consider both grammar scale and adaptation characteristics. Systematic and repetitive adaptations favor rule-based methods for their deterministic execution, while context-dependent adaptations that vary across rules benefit from LLM-based approaches' reduced configuration effort. For grammars between 100-200 rules, practitioners should evaluate whether adaptation complexity outweighs the risks of systematic omission, potentially testing both approaches on representative subsets.}
% \todo[inline]{Weixing: The content here should be updated accordingly, because something has been changed.}
\subsection{Threats to Validity}
\paragraph{Internal Validity:} Several factors in our study may affect the validity of causal inferences. First, we used only Claude Sonnet 4.5 for prompt strategy development, then applied the finalized prompts to ChatGPT 5.1 and Gemini 3 for cross-model validation; while this approach evaluates the cross-model consistency of prompts, using a different reference model for prompt development might yield different prompt strategies and performance characteristics. 
% Second, the inherent non-determinism of LLMs poses a threat to the reliability of results: each of our experiments was a single run, without evaluating the output stability of the same prompt across multiple executions. 
Second, the decision criteria for prompt finalization (>90\% RAC threshold) were chosen empirically; 
% although this threshold did not affect the final prompt selection on our training set, 
under different training sets or DSL characteristics, different thresholds might lead to different prompt strategies, thereby affecting performance on the test set.
\revised{Third, the finalized prompts met all quality thresholds on the small-to-medium scale training DSLs without iterative refinement.
% , suggesting that a general-purpose prompt suffices at this scale but may not generalize to scenarios requiring more complex prompting strategies.
% Furthermore, we did not compare our general-purpose prompts against alternative prompting strategies such as few-shot or chain-of-thought prompting, which remains a direction for future work.
}

% Construct Validity: Evaluating whether LLM-generated adaptations 
% follow the same patterns as manual adaptations requires subjective 
% judgment. Future work could develop formal metrics or automated 
% tools to more objectively assess adaptation similarity.

\paragraph{Construct Validity:}Our evaluation metrics—RAC, Output Similarity, and Metamodel Conformance—while providing objective and quantifiable measurements, may not fully capture the multidimensional nature of ``high-quality grammar adaptation''. RAC measures the consistency of adaptations, i.e., whether adaptations are correctly reused, but 
does not evaluate the reasonableness or desirability 
of the adaptations themselves. For example, if the target grammar ($G_1'$) itself contains design flaws or does not conform to best practices, the LLM's perfect replication of these adaptations would still achieve 100\% RAC, yet the quality of the resulting grammar may not be ideal. Furthermore, our metrics also focus on metamodel conformance and grammar rule similarity, but they do not evaluate other quality properties that grammar engineers might care about, e.g., grammar readability, maintainability, consistent naming conventions, or adherence to team coding style guidelines~\cite{dubey2006goodness}. 
Therefore, while our metrics can effectively validate whether LLMs learn and apply historical adaptations, they may not be sufficient to comprehensively assess the overall quality of adapted grammars in practical software engineering practice.

\paragraph{External Validity:}
% Several factors limit the generalizability of our findings. First, we evaluated our approach on only two DSLs (QVTo and EAST-ADL) with a limited number of versions. While these represent realistic metamodel-grammar co-evolution scenarios, the results may not generalize to other DSLs with different characteristics or evolution patterns. 
Although we evaluated our approach on six DSLs from different domains, the sample size remains limited, making it difficult to draw universally applicable conclusions. Furthermore, in the evaluation of these six DSLs, the $G_1$ and $G_2$ we used are actually the same generated grammar. However, in actual language evolution scenarios, metamodel changes may lead to differences between the generated grammar $G_2$ and $G_1$, e.g., adding or deleting grammar rules. 
% This simplified experimental setup may not adequately reflect the complexity of real evolution contexts, posing a threat to external validity. 
Our longitudinal evaluation on QVTo mitigates this threat to some extent, as QVTo's four official versions did undergo real metamodel evolution, with actual differences in the generated grammars between versions. 
However, the changes between QVTo versions are relatively small as shown in Table~\ref{tab:qvto_evolution}, which may not capture the full complexity of real evolution scenarios.
% sustained language evolution processes.
Second, our approach is specifically designed for Xtext-based DSLs, where grammars can be generated from metamodels and the generated grammars follow Xtext's conventions and patterns. The applicability of this approach to DSLs developed with other frameworks (such as JetBrains MPS~\cite{jetbrainmps}) remains unclear.
% \updated{Third, although our DSLs exhibit domain diversity (from automotive engineering's EAST-ADL to bibliography management's BibTeX), their coverage in terms of grammar adaptations may be limited. The adaptations we evaluated primarily involve grammar rule-level modifications, e.g., keyword adjustments, attribute reorganization, optionality changes, delimiter modifications, etc., which may not represent all grammar evolution scenarios that might occur in practice. Similarly, the ``complex scenarios'' we encountered in DOT and Xcore (such as syntactic predicates and order-insensitive combinations), while challenging for rule-based approaches, may not represent all types of grammar adaptation complexity.}
Third, \revised{we intentionally focused on a setup decoupled from specific DSL examples rather than few-shot prompting, as example-based prompting can steer outputs toward the provided examples and limit generalizability; comparison against alternative strategies such as chain-of-thought prompting remains a direction for future work.
Furthermore,} our approach encountered significant challenges on the large-scale grammar (i.e., the grammar of EAST-ADL), forcing us to exclude EAST-ADL from the prompt finalization process. This indicates that grammar scale may be a critical factor affecting the effectiveness of LLM-based approaches, and the small-to-medium scale DSLs on which our final evaluation is based may not represent the actual challenges of large-scale industrial DSLs. Future research needs to explore strategies specifically designed for large-scale grammars, such as segmented processing, incremental adaptation, etc., to overcome the systematic omission problems of current LLMs when handling hundreds of grammar rules.

% Conclusion Validity: Evaluating whether LLM-generated adaptations 
% follow the same patterns as manual adaptations requires subjective 
% judgment. Future work could develop formal metrics or automated 
% tools to more objectively assess adaptation similarity.
\subsection{The Role of LLMs in Model-Driven Engineering}
This study explores the potential of LLMs in automating grammar adaptation tasks, contributing empirical evidence to the intersection of AI and MBE. This intersection manifests in two complementary directions: using LLMs to support MDE tasks (LLM4MDE), and using MDE techniques to facilitate the adoption of LLMs (MDE4LLM)~\cite{di2025use}. %Our work belongs to the former.
% , leveraging LLMs' capabilities to automate adaptations during grammar evolution processes.

\paragraph{LLMs as Components of the MBE Toolchain.} LLMs as Components of the MBE Toolchain. Unlike traditional rule-based MDE tools, LLMs introduce probabilistic and non-deterministic characteristics~\cite{astekin2024exploratory, sawadogo2025revisiting, ouyang2025empirical}, posing challenges for MDE toolchains that traditionally rely on deterministic transformations to ensure model consistency and traceability~\cite{lucas2009systematic, bucchiarone2020grand}. 
However, the ``black box'' nature of LLMs means we cannot fully predict or explain how their generated outputs are produced. In our study, this characteristic is partially mitigated through mandatory metamodel conformance validation—any grammar generated by LLMs must pass validation through the Eclipse Modeling Framework, ensuring its semantic correctness. 
% This pattern of combining probabilistic AI methods with formal verification may represent a viable architecture for future AI-enhanced MDE tools: leveraging the flexibility and pattern learning capabilities of LLMs while providing necessary guarantees through traditional formal methods.
This pattern of combining probabilistic AI methods with formal verification may represent a viable architecture for future AI-enhanced MDE tools.

\paragraph{Specificity of Prompt Engineering in MDE Contexts.} Our prompt development process reveals the unique requirements of prompt engineering in the MDE domain. Unlike general LLM applications, MDE tasks have clear correctness criteria—generated grammars must conform to metamodel specifications. This enables us to adopt more stringent quality thresholds than traditional prompt engineering: prompts are finalized only when LLMs achieve (RAC>90\%), and output (similarity>90\%), and pass metamodel conformance validation on the training set. 
% The uniqueness of this prompt evaluation approach lies in that it not only uses objective quantitative metrics (RAC, output similarity), but these metrics can be directly computed against pre-existing target grammars without requiring additional manual annotation; more importantly, we 
Notably, these metrics can be computed directly against pre-existing target grammars without manual annotation, and metamodel conformance validation provides a formal correctness guarantee, ensuring the semantic validity of generated grammars. 
Furthermore, our training-test separation methodology—using four DSLs for prompt development and two DSLs for testing—reflects scientific rigor aimed at avoiding prompt overfitting to specific cases.
% Furthermore, our training-test separation methodology—using four DSLs for prompt development and two DSLs for testing—reflects our commitment to scientific rigor, aiming to avoid prompt overfitting to specific cases. 
% These practices provide methodological references for how to systematically engineer LLM applications in domains with formal requirements (such as MDE).

\paragraph{From Tool Development to Methodological Exploration.} The value of this study lies not only in demonstrating that LLMs can perform grammar adaptation tasks, but more importantly in systematically characterizing the capability boundaries of this approach. Through evaluation on DSLs of different scales and complexities, we identified the applicable domain of LLM methods (small-to-medium scale, complex adaptations requiring context-dependent understanding) and the limitation domain (large-scale, requiring systematic application of hundreds of operations). 
% This understanding of capability boundaries is crucial for MDE practitioners when selecting automation strategies.
% and also points to directions requiring further exploration by the research community. For example, how to design hybrid approaches that combine LLMs' contextual understanding capabilities with the reliability and scalability of rule-based methods? How to overcome LLMs' systematic omission problems in large-scale grammar processing? Exploration of these questions will help advance the development of AI-enhanced MDE tools. 
\revised{It also suggests a concrete hybrid architecture: since the identification stage was not the primary bottleneck for Claude and Gemini on EAST-ADL while the application stage exhibited systematic omissions, one direction would be to use LLMs exclusively for identifying adaptation operations from $G_1$ and $G_1'$, and then delegate execution to a deterministic rule-based engine such as GrammarTransformer. However, realizing this architecture faces a challenge: LLMs produce informal natural language adaptation descriptions (e.g., ``move shortName outside curly braces'') rather than formal, executable rule specifications, requiring either constrained output formats or a translation layer to the 60 predefined transformation rules. 
% Bridging this gap would require either constraining LLM outputs to a structured format or developing a translation layer between natural language descriptions and the 60 predefined transformation rules. 
Moreover, some adaptations that LLMs successfully handle (such as syntactic predicates and order-insensitive attribute combinations) fall outside the coverage of current predefined rules, meaning that the rule-based execution engine would also need to be extended.}
In this sense, our study provides systematic empirical analysis of ``the role of AI methods in grammar adaptation as a specific MDE task,'' complementing existing literature's exploration of LLMs in other MDE tasks (such as model generation and model completion~\cite{arulmohan2023extracting, chen2023automated}).

\paragraph{Enhancing Metamodel-Driven Development Viability.}
Prior large-scale study~\cite{zhang2024tales, zhang2026development} revealed that Xtext-based DSL development predominantly follows the grammar-driven scenario, with metamodel-driven approaches being less common. A reason for this preference is the manual workload: 
In scenario (b), after metamodel evolution, developers regenerate $G_2$ and reapply all previous adaptations, whereas scenario (a) requires only localized adjustments to $G_1'$ based on metamodel changes.
% . In contrast, scenario (a) requires only localized adjustments to $G_1'$ based on metamodel changes.
% While the existing rule-based method partially alleviates this issue, 
our study demonstrates that for small-to-medium scale DSLs, LLM-based approaches can further automate the learning and reapplication of historical adaptations, reducing this workload and making scenario (b) more attractive relative to scenario (a).

\subsection{LLM-based Approach for Grammar Style Customization}
% \todo[inline]{Weixing: need a double check}
Beyond supporting metamodel-grammar co-evolution, our LLM-based approach opens new possibilities for automated grammar style customization. Previous work demonstrated a semi-automated approach for transforming Xtext-generated grammars into specific styles, such as Python-style DSLs~\cite{zhang2023creating}, requiring manual keyword refinement decisions and predefined transformation scripts for operations like brace removal and whitespace-awareness introduction.

% The findings of this study suggest the potential application value of LLMs in grammar style customization. 
Many of the adaptation operations that LLMs successfully handled in this study—such as removing braces, refining keywords, and adjusting attribute positions—are similar in nature to those in prior style customization work. 
\section{Related Work}
\label{sec:relatedwork}
% \todo[inline]{Assigned to @Rahul}

\subsection{Metamodel-Grammar Co-evolution in MDE}
The evolution of DSLs is a %well-documented 
challenge in MDE. As detailed by Meyers et al.\cite{MEYERS20111223}, the "coupled evolution" problem arises when changes in a metamodel (abstract syntax) necessitate the propagation of updates to related artifacts, such as models, transformations, and grammars (concrete syntax). Traditionally, this process has been handled through manual adaptation or rule-based automation. For instance, in the context of Xtext,changes to the Ecore metamodel often break the mapping to the Xtext grammar. 

Zhang et al.\cite{zhang2024supporting} addressed this by proposing GrammarTransformer, an approach that utilizes predefined transformation rules to semi-automate the synchronization between evolved metamodels and grammars. 

Similarly, Kusel et al.\cite{Kusel2015} provided a systematic classification of co-evolution approaches, highlighting that while operator-based and inference-based methods exist, they often require significant manual configuration or rigorous formal definitions that are difficult to maintain in rapid development cycles. Our work differs from these traditional approaches by replacing rigid transformation rules with the probabilistic reasoning capabilities of LLMs, aiming to learn adaptations implicitly from history rather than requiring explicit rule definition.

\subsection{Large Language Models in Software Engineering}
% The advent of LLMs has fundamentally shifted the landscape of automated software engineering. 
\updated{The emergence of Large Language Models (LLMs) has significantly impacted numerous domains within Software Engineering (SE). Many recent studies have explored applying LLMs to various SE tasks
% , aiming to optimize processes and outcomes
~\cite{hou2024large}.}
% Models such as OpenAI's GPT-4 and Meta's LLaMA 
LLMs
have demonstrated state-of-the-art performance in code generation, completion, and translation~\cite{chen2021}.
% \updated{and recent surveys show their effectiveness extends to software testing tasks, with Wang et al.~\cite{wang2024software} analyzing 102 studies spanning test case generation, program debugging, and program repair.}

In the specific domain of MDE, the integration of LLMs is gaining traction~\cite{di2025use}. 
\updated{Chaaben et al.~\cite{chaaben2023towards} pioneered the use of few-shot prompting for model completion, demonstrating that GPT-3 can complete UML models when provided with examples from the ModelSet dataset.}
Studies have explored using LLMs to generate OCL constraints~\cite{pan2024generative}, reverse engineer models from code~\cite{pearce2022pop}, and assist in creating DSL instances~\cite{netz2024natural}. For example, LLMs can effectively parse natural language requirements into formal model specifications, reducing the barrier to entry for DSL users. However, these applications generally focus on generation (creating new artifacts) rather than evolution (maintaining consistency between existing, changing artifacts), which requires a deeper understanding of the structural dependencies inherent in DSLs.

\subsection{LLM-based Artifact Co-evolution}
Most relevant to our study is the emerging body of work applying LLMs specifically to the co-evolution and migration of software artifacts. Jiang et al.\cite{Jiang2025} introduced Codeditor, an LLM-based tool designed to migrate code changes across programming languages (e.g., Java to C\#), demonstrating that LLMs can learn edit patterns more effectively than traditional transpilers. 
% Closer to the MDE context, Kebaili et al. \cite{Kebaili2024} conducted an empirical study on using ChatGPT to co-evolve source code in response to metamodel changes. Their results indicated that while LLMs could successfully repair a significant portion of broken code (achieving up to 88.7\% correctness in specific scenarios), the approaches were heavily dependent on prompt engineering and did not specifically target the grammar definition itself. 
\updated{Closer to the MDE context, Kebaili et al.~\cite{Kebaili2024} conducted an empirical study using ChatGPT to co-evolve source code in response to metamodel changes. Their results showed that while LLMs can successfully repair most broken code (achieving up to 88.7\% correctness in specific scenarios), %these approaches heavily rely on prompt engineering and 
specifically target generated code repair. In contrast, our work targets the adaptation of the grammar definition itself—a more upstream challenge in the language infrastructure.}
Furthermore, recent investigations by Zhang et al.~\cite{zhang2025leveraging, zhang2026leveraging} have begun to explore the co-evolution of textual DSL instances using LLMs, evaluating how models can migrate user scripts when the underlying grammar changes. Our work complements and extends these efforts by shifting the focus upstream: instead of fixing instances or generated code.
% , we target the grammar definition itself. 
By evaluating LLMs' ability to adapt Xtext grammars directly from metamodel changes—demonstrated across real-world DSLs like QVTo.
% —we address a critical gap in automating the maintenance of the language infrastructure itself.

\section{Conclusion}
\label{sec:conclusion}
% This paper explores the potential of large language models in automating metamodel-grammar co-evolution. We propose an LLM-based approach that automatically adapts grammars by learning from historical adaptations, without relying on predefined transformation rules. Evaluation on six real-world DSLs shows that three mainstream LLMs achieved 100\% \revised{grammar rule-level} adaptation consistency and output similarity on the test set (DOT, Xcore), while the rule-based approach achieved 84.21\% and 87.50\% on DOT, and 62.50\% and 70.00\% on Xcore, respectively. The QVTo longitudinal study demonstrates that the LLM-based approach 
% % achieved 
% \revised{successfully reused learned adaptations across all three evolution steps without manual grammar editing}
% % zero human intervention 
% across three evolution steps, while the rule-based approach required configuration adjustments in two transitions. Results indicate that LLMs excel at handling complex grammar scenarios (such as syntactic predicates, predicated assignments, and order-insensitive attribute combinations), but encounter systematic omission of identified adaptation operations on large-scale grammars (such as EAST-ADL with 297 rules), with RAC values far below 90\%.
This paper explores the potential of large language models in automating metamodel-grammar co-evolution. We proposed an LLM-based approach that learns grammar adaptations from historical versions and applies them to new grammars after metamodel evolution. Evaluation on six real-world DSLs demonstrates that LLMs outperform the rule-based approach on complex grammar scenarios and successfully reuse adaptations across consecutive evolution steps without manual grammar editing; however, systematic omission of adaptation operations on large-scale grammars reveals the current limitations of LLMs at scale.

% This study demonstrates new possibilities for integrating AI with traditional model-driven approaches and provides a replicable evaluation framework. Future work includes three directions: improving LLM reliability on large-scale grammars, extending to co-evolution of other language artifacts (such as model transformation rules and OCL constraints with metamodels), and exploring the potential of LLMs in supporting co-evolution of modeling artifacts at different abstraction levels.
Future work includes three directions: improving LLM reliability on large-scale grammars, extending to co-evolution of other language artifacts (such as model transformation rules and OCL constraints with metamodels), and exploring the potential of LLMs in supporting co-evolution of modeling artifacts at different abstraction levels.

\bibliography{main}

\section*{About the authors}
\shortbio{Weixing Zhang}{is a Postdoctoral researcher at Karlsruhe Institute of Technology. \authorcontact[https://wilson008.github.io/]{weixing.zhang@kit.edu}}

\shortbio{Bowen Jiang}{is a PhD researcher at Karlsruhe Institute of Technology. \authorcontact[https://mcse.kastel.kit.edu/staff_bowen_jiang.php]{bowen.jiang@kit.edu}}

\shortbio{Rahul Sharma}{is a Postdoctoral researcher at Karlsruhe Institute of Technology. \authorcontact[https://dsis.kastel.kit.edu/staff_rahul_sharma.php]{rahul.sharma@kit.edu}}

\shortbio{Ragina Hebig}{is a Professor for Software Engineering at the University of Rostock, Germany.  \authorcontact[https://se.informatik.uni-rostock.de/team/lehrstuhlinhaber/prof-dr-rer-nat-regina-hebig/]{regina.hebig@uni-rostock.de}}

\shortbio{Daniel Strüber}{is an Asso. Professor at Chalmers $|$ University of Gothenburg, Sweden, and an Ass. Professor at Radboud University in Nijmegen, the Netherlands. \authorcontact[https://www.danielstrueber.de/]{danstru@chalmers.se}}

%\shortbio{Rahul Sharma}{is a Postdoctoral researcher at Karlsruhe Institute of Technology. \authorcontact[https://dsis.kastel.kit.edu/staff_rahul_sharma.php]{rahul.sharma@kit.edu}}

%\shortbio{Bowen Jiang}{is a PhD researcher at Karlsruhe Institute of Technology. \authorcontact[https://mcse.kastel.kit.edu/staff_bowen_jiang.php]{bowen.jiang@kit.edu}}

\end{document}